\documentclass[a4paper, 10pt, conference]{ieeeconf}      

\IEEEoverridecommandlockouts                              

\usepackage{graphicx}
\usepackage{amsmath} 
\usepackage{amssymb}  
\usepackage{url}  
\usepackage{color}  
\usepackage{subcaption}
\usepackage{acronym}
\usepackage{hyperref} 

\newacro{fov}[FOV]{field-of-view}

\newcommand{\transform}[3]{
^{#2}#1_{#3}
}

\title{\LARGE \bf
A Flexible Framework for Virtual Omnidirectional Vision to Improve Operator Situation Awareness
}

\author{Martin Oehler and Oskar von Stryk
\thanks{All authors are with the Simulation, Systems Optimization and Robotics Group, Technical University of Darmstadt, Germany
        \mbox{\tt\small \{oehler, stryk\}@sim.tu-darmstadt.de}
}%
\thanks{Research presented in this paper has been supported in parts by the German Federal Ministry of Education and Research (BMBF) within the subproject "Autonomous Assistance Functions for Ground Robots" of the collaborative A-DRZ project (grant no. 13N14861), by the German Federal Ministry of Economics \& Technology’s "EXIST Forschungstransfer" (grant no. 03EFLHE061) and by Nexplore within the AICO Collaboration Lab. This work has been co-funded by the LOEWE initiative (Hesse, Germany) within the emergenCITY center.
\newline 978-1-6654-1213-1/21/\$31.00 \textcopyright 2021 IEEE}%
}

\AtBeginShipoutFirst{\raisebox{0cm}{\parbox{0.9\textwidth}{\centering\large Preprint of the paper which appeared in:\\
\LARGE European Conference on Mobile Robots (ECMR) 2021}}}

\begin{document}

\maketitle
\thispagestyle{empty}
\pagestyle{empty}

\begin{abstract}
During teleoperation of a mobile robot, providing good operator situation awareness is a major concern as a single mistake can lead to mission failure. Camera streams are widely used for teleoperation but offer limited \acl{fov}.
In this paper, we present a flexible framework for virtual projections to increase situation awareness based on a novel method to fuse multiple cameras mounted anywhere on the robot.
Moreover, we propose a complementary approach to improve scene understanding by fusing camera images and geometric 3D Lidar data to obtain a colorized point cloud.
The implementation on a compact omnidirectional camera reduces system complexity considerably and solves multiple use-cases on a much smaller footprint compared to traditional approaches such as actuated pan-tilt units.
Finally, we demonstrate the generality of the approach by application to the multi-camera system of the Boston Dynamics Spot.
The software implementation is available as open-source ROS packages on the project page \url{https://tu-darmstadt-ros-pkg.github.io/omnidirectional_vision}.

\end{abstract}
\section{INTRODUCTION}

Teleoperated robots are often deployed for tasks in hazardous environments like search and rescue. They enable the operator to safely explore and interact with the environment by remote operation.
However, operator errors
are a common reason for the failure of a robot mission.
A frequent cause for these errors is a lack of situation awareness \cite{norton2017analysis}.

Commonly, remotely operated robots are equipped with a camera to stream video data to the operator. Compared to the human field-of-view, standard perspective cameras are very limited and lead to ``tunnel vision''. The peripheral vision loss hinders the operator's ability to build a mental model of the robot's surroundings.

In many cases, this problem is addressed by one or a combination of the following approaches: 1) Mounting a wide-angle fish-eye camera instead of a standard camera \cite{chen2007human} increases \ac{fov} in one direction. The severe distortion of these images makes the judging of spatial relations very challenging for the operator. Furthermore, blind spots remain to the sides and behind the vehicle.
2) Additional cameras can be mounted to cover these blind spots at the cost of increased system complexity. This increases the mental load of the operator because attention has to be switched between views~\cite{chen2007human}. In practice, operators mainly focus on the primary camera and often miss changes in auxiliary views \cite{keyes2006camera}.
3) A wide-spread alternative is a mechanical pan-tilt sensor head that points a mounted camera in the desired viewing direction. These systems are complex with higher space requirements compared to individual cameras, limiting applicability on small robot platforms. Further drawbacks are introduced by the actuation. Mechanical limits can lead to blind spots and slow-moving actuators and network latency can result in a low control responsiveness.

A further problem of teleoperation using a video stream is the lack of depth information. As a result, estimating the distance to obstacles is challenging, making navigation in narrow spaces demanding and error-prone \cite{chen2007human}.
There are different sensor modalities available that provide depth information, with Lidar and depth cameras being the most popular choices. Lidar sensing provides high-precision range information with a high field-of-view, but a fundamental drawback is the lack of color information. Based on the geometric range data alone, it is challenging for humans to understand the environment. For instance, doors are hard to distinguish from walls. Most depth cameras provide color information, but their range, accuracy and \ac{fov} are limited compared to Lidar.

\begin{figure}[tbp]
    \centering
    \includegraphics[width=\linewidth]{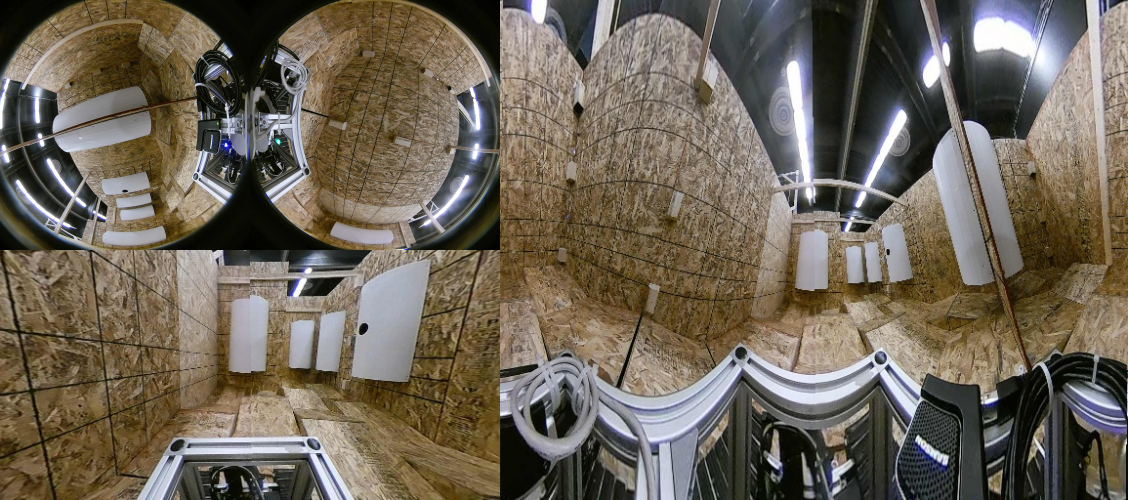}
    \caption{Two fish-eye images are streamed from the omnidirectional camera (top-left), which are converted to perspective (bottom-left) and Mercator projection (right).}
    \label{fig:image_projection_preview}
\end{figure}

In this paper, we propose two complementary approaches improving operator situation awareness using omnidirectional vision.
We present a flexible framework to compute virtual projections from multiple cameras mounted anywhere on the robot. This allows generating novel views, giving the operator an unrestricted 360° view of the environment (see Fig.~\ref{fig:image_projection_preview}). By recomputing the projection online, arbitrary frames on the robot or in the environment can be tracked. In this way, a virtual pan-tilt sensor head can be realized.
This is complemented by a second approach that improves scene understanding by fusing Lidar data with color information from multiple cameras simultaneously to obtain a colorized point cloud.

The combination of both approaches in a comprehensive user interface allows for an improved situation-aware teleoperation. By generating virtual views, the operator can navigate with rectified pinhole projections, which are easier to comprehend than distorted fish-eye images. Spatial awareness is increased by additionally displaying a colorized point cloud in a 3D scene.
The implementation of both approaches is made available as open-source ROS packages\footnote{\url{https://tu-darmstadt-ros-pkg.github.io/omnidirectional_vision}}.

\begin{figure}[tbp]
    \begin{subfigure}[t]{0.58\linewidth}
        \includegraphics[width=\linewidth]{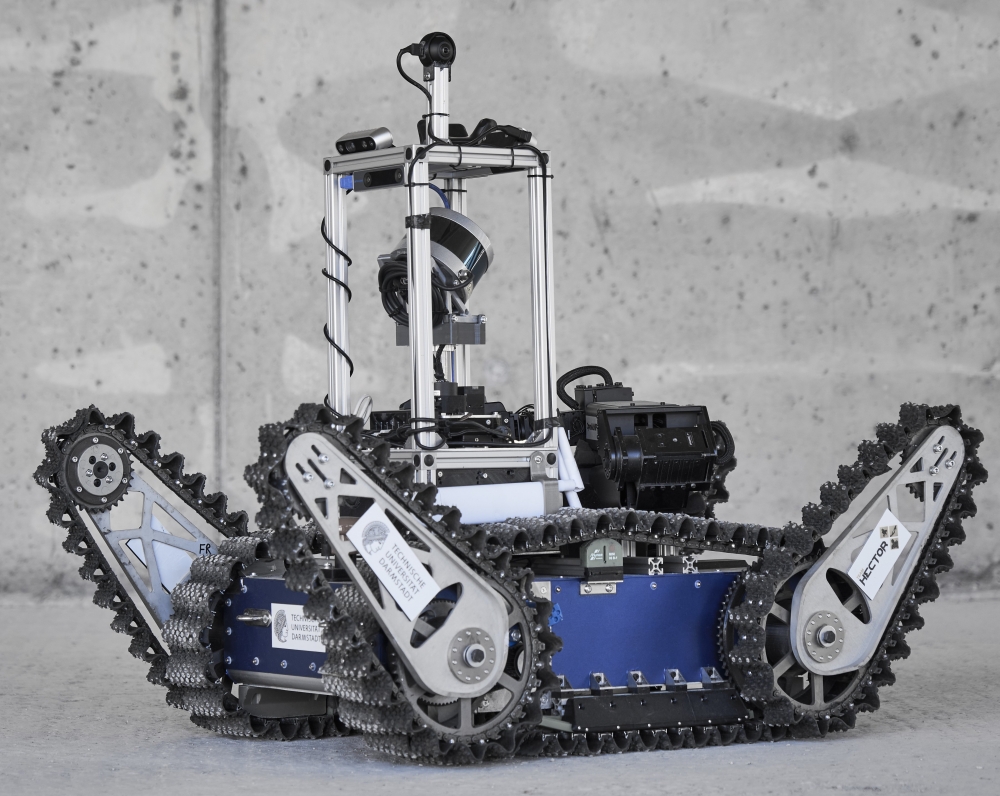}
        \caption{\textit{Asterix}}
        \label{fig:asterix}
    \end{subfigure}
    \begin{subfigure}[t]{0.4\linewidth}
    \includegraphics[width=\linewidth]{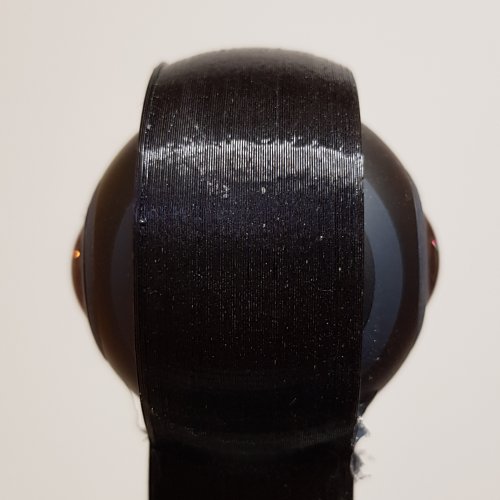}
    \caption{Insta360 Air\label{fig:insta360_air}}
    \end{subfigure}
    \caption{The \textit{Insta360 Air} mounted on top of the center-frame of the highly mobile tracked platform \textit{Asterix}.}
\end{figure}

We demonstrate advantages compared to traditional approaches on two different multi-vision systems. First, we implement our approach on the \textit{Insta360 Air} (Fig. \ref{fig:insta360_air}) mounted on the tracked robot \textit{Asterix}~\cite{schnaubelt_mechatronik_2021} (Fig.~\ref{fig:asterix}). While we used a complex multi-camera system in the past, switching to a single compact omnidirectional camera reduced system complexity considerably and solved different use-cases on a much smaller footprint. We highlight the generality of our projection framework by implementing it on the multi-camera system of the \textit{Boston Dynamics Spot}. With five standard cameras mounted on the robot body, its camera configuration is significantly different to the \textit{Insta360 Air}. \textit{Spot} in its standard configuration does not feature a directly forward-facing camera. Therefore, we fuse the two angled front cameras to generate a view suitable for teleoperation. As a quadrupedal system, Spot can move in any direction. In this case, it is especially important to be aware of obstacles to the sides. To improve awareness of surrounding obstacles, we generate a virtual bird's eye view that shows the environment from above.

\section{RELATED WORK}
Various approaches to improve operator situation awareness using omnidirectional vision have been proposed.

The authors of \cite{keyes2006camera} compare the effect of different camera placements on the teleoperation performance. They find that visualizing the robot and its surroundings in one image significantly reduces collisions. This view can be achieved by an overhead camera, but mounting a pole for attachment is impractical.
In \cite{sato2013spatio}, the authors compute a virtual bird's-eye view from multiple fish-eye cameras by transforming the images to the ground plane with a homography. This leads to a distortion of obstacles, because their height does not match the ground.
This problem is addressed in \cite{awashima2017visualization} by incorporating data from depth sensors. Obstacles are superimposed on the bird's-eye view to correctly visualize their location.

An overview of human performance issues related to user interface design is given in \cite{chen2007human}. A low \ac{fov} is identified to cause erroneous speed and distance judgments as well as peripheral vision loss.
In \cite{vaughan2016look}, a fish-eye camera is used to increase operator \ac{fov}. The authors present an approach to generate a virtual focus and context view that shows a perspective view in the center and transitions to a distorted view in border regions. This allows for an undistorted view in the center while still providing peripheral information.

The authors of \cite{norton2017analysis} compare different approaches for human-robot interaction in a case study of the DARPA Robotics Challenge Finals. They conclude that situation awareness is improved when data is visualized in a virtual environment and the operator is able to control the viewpoint. In \cite{iwataki2015visualization}, this is achieved by projecting the image of four fish-eye cameras onto a hemispherical mesh.
The approach leads to distortions because the mesh does not match the environment. Moreover, scale depends on the mesh size, which can be irritating.
This is improved in \cite{komatsu2020free} by taking the environment into account. An environment mesh is approximated with laser range finders and textured with images from multiple fish-eye cameras. While this provides good situation awareness in certain environments, the assumption that only flat walls are present leads to a distortion of obstacles.
The authors of \cite{vunder2018improved} take a similar approach and project the images of two fish-eye cameras to a sphere. They render perspective views by placing a virtual camera inside the sphere and display this image on a head-mounted display with orientation tracking for an intuitive visualization. As part of our evaluation, we compare this approach to ours.
Additionally, they propose a second approach to project Lidar points onto the sphere to fuse color information with depth.
Further research on colorizing point clouds has been performed in \cite{vechersky2018colourising}. The authors propose an approach to color Lidar point data with an attached camera. They fuse color information from multiple observations and take visibility and temporal offsets into account.

\section{METHOD}
In the following section, we describe our approach to create projections from multiple source cameras. Afterwards, we explain how point cloud data is colorized from the same cameras.

\subsection{Camera and Lidar Calibration}

The described approaches require an accurate calibration of intrinsic and extrinsic parameters of all cameras and the extrinsic calibration of the Lidar.

The general model of projection for a single camera $C_i$ is given by:

\begin{equation}
    \mathbf{\pi}_{c_i}\left(^{C_i}T_R \mathbf{p}\right) = \mathbf{u}
\end{equation}

The extrinsic camera calibration describes the relative 6D pose of each camera to a common reference frame $R$, e.g. the first camera or a common camera head. It is given by the SE(3) transformation $\transform{T}{R}{C_i}$. The inverse $\transform{T}{C_i}{R}$ transforms point $\mathbf{p} = \left(x, y, z\right)$ from reference frame to camera frame. The intrinsic camera calibration yields the projection function $\mathbf{\pi}_{c_i}: \mathbb{R}^3 \rightarrow \mathbb{R}^2$ which projects a 3D point onto a 2D image coordinate $\mathbf{u} = \left(u, v\right)$. Note that we use a general projection model and no assumption about a specific lens type is made. This allows our approach to generalize to different projection models, e.g. the standard pinhole camera model and wide-angle lens models. Extrinsic and intrinsic calibration are performed jointly for all cameras with the calibration framework \textit{kalibr}\footnote{\url{https://github.com/ethz-asl/kalibr}}~\cite{furgale2013unified}.

In a second step, the extrinsic calibration between camera and Lidar is estimated. It describes the 6D pose of the Lidar frame $L$ relative to the camera reference frame $R$ given by the transform $\transform{T}{R}{L}$. A target-less calibration is performed using the approach described in \cite{pandey2012automatic}.

\subsection{Image Projection with Multiple Cameras}

A projection is defined by a function $P: \mathbb{R}^2 \rightarrow \mathbb{R}^2$ which maps coordinates from the output projection image to the input camera image. In case of multiple cameras, one projection function is defined per camera:
\begin{equation}
    P_{c_i}(\mathbf{u}_p) = \mathbf{u}_{c_i}
\end{equation}
with $\mathbf{u}_p = (u_p, v_p) $ being the projection image coordinates and $\mathbf{u}_{c_i} = (u_{c_i}, v_{c_i})$ the image coordinates in camera ${C_i}$ respectively. The projection function is used to look up camera image coordinates for each pixel of the output image.

Geometrically, projections can be represented by 3D surfaces.
The shape of the surface depends on the projection type and parameters, which should be chosen based on the desired application.
The 6D pose of the projection surface frame $P$ relative to the camera reference frame $R$ is specified by the user with the transform $\transform{T}{R}{P}$.

Based on the projection surface, the projection function is computed. First, the surface is discretized by grid-based sampling with the desired target resolution $\left(W_p \times H_p\right)$. Each 3D sample point $\mathbf{x}_p = (x_p, y_p, z_p)$ corresponds to a pixel in the output image $(u_p, v_p)$. This mapping between 2D projection image and 3D projection surface is given by the surface projection function $S: \mathbb{R}^2 \rightarrow \mathbb{R}^3$:

\begin{equation}
    S(u_p, v_p) = \mathbf{x}_p = \begin{pmatrix}
                                  x_p \\
                                  y_p \\
                                  z_p
                                  \end{pmatrix}
\end{equation}

The projection surface function for three different projections is detailed in the following.

\emph{Perspective projection:} The perspective projection surface function describes a plane. The parameters focal length $f$ and vertical \ac{fov} $\phi_h$ of the pinhole model are specified by the user. Based on this, the pixel size $m_p$ is computed:
\begin{equation}
    m_p = \frac{2 f \tan\left(\phi_h/2\right)}{W_p}
\end{equation}
which leads to:
\begin{align}
    S\left(u_p, v_p\right) =  \begin{pmatrix}
                                (u_p - W_p / 2) \cdot m_p \\
                                (v_p - H_p / 2) \cdot m_p \\
                                f
                                \end{pmatrix}
\end{align}

\emph{Mercator projection:} The Mercator projection is represented by a cylinder and gives a panoramic view of the environment. The desired vertical \ac{fov} $\phi_v$  and cylinder radius $c_r$ are specified by the user. Based on this, angular pixel size $\alpha_p$ and cylinder height $c_h$ are computed:

\begin{align}
    \alpha_p = \frac{2\pi}{W_p} \\
    c_h = 2c_r \tan\left(\phi_v\right)
\end{align}

The surface projection function is then given by:
\begin{align}
    S\left(u_p, v_p\right) =  \begin{pmatrix}
                                c_r \cos(-u_p \alpha_p) \\
                                c_r \sin(-v_p \alpha_p) \\
                                c_h (0.5 - \frac{v_p}{H_p})
                                \end{pmatrix}
\end{align}

\emph{Spherical projection:} A spherical projection can be geometrically represented by a sphere. The \ac{fov} $\phi$ and sphere radius $s_r$ are given by the user. First, the projection image coordinates $(u_p, v_p)$ are transformed to normalized polar coordinates $r \in [0;1]$ and $\gamma \in (-\pi, \pi]$. The origin of the polar coordinate system is located in the center of the image:

\begin{align}
    p_x &= \frac{u_p}{W_p} - 0.5 \\
    p_y &= 0.5 - \frac{v_p}{H_p} \\
    r &= \sqrt{p_x^2 + p_y^2} \\
    \gamma &= \arctan2(p_y, p_x)
\end{align}

To cast a ray into 3D, we also compute the second angle $\theta$ which corresponds to the latitude while $\gamma$ corresponds to the longitude of the sphere:
\begin{equation} \label{eq:rel_lat_r}
    \theta = r \frac{\phi}{2}
\end{equation}

Based on latitude $\theta$ and longitude $\gamma$, the sphere is computed:
\begin{align}
    S\left(\theta, \gamma \right) =  \begin{pmatrix}
                                s_r  \sin{\theta}  \cos{\gamma} \\
                                -s_r  \sin{\theta}  \sin{\gamma} \\
                                s_r  \cos{\theta}
                                \end{pmatrix}
\end{align}
Fish-eye lenses produce spherical images. An ideal fish-eye lens has a linear relationship between distance from image center $r$ and latitude $\theta$ as can be seen in (\ref{eq:rel_lat_r}). Real fish-eye lenses are rarely linear, therefore, a projection can be used to "rectify" the image.

Using the surface projection function $S$, a surface sampling point $\mathbf{x}_p$ is computed for each pixel in the output image.

In the next step, each point $\mathbf{x}_p$ is transformed into the camera frame of each camera using the user-specified surface pose $\transform{T}{R}{P}$ and the extrinsic camera calibration $\transform{T}{C_i}{R}$:

\begin{equation}
    \mathbf{x}_{c_i} = \transform{T}{C_i}{R} \,\transform{T}{R}{P} \mathbf{x}_p
\end{equation}

Finally, we project the point into all camera images with the intrinsic calibration function $\mathbf{\pi}(\cdot)$ of each camera:

\begin{equation} \label{eq:camera_projection}
\mathbf{\pi}_i\left( \mathbf{x}_{c_i} \right) = \mathbf{u}_{c_i}
\end{equation}

Using this process, we compute the mapping $P_{c_i}$ for each pixel of the output image $\mathbf{u}_p$ and for each camera $C_i$. In the case where we have multiple valid mappings for a pixel, i.e. when the projection surface is seen by multiple cameras, we prefer the mapping, where $\mathbf{u}_{c_i}$ is closest to the respective image center. Alternatively, one could consider all observations by blending their color.

The projection mapping function has to be computed only once. Each time a new image is received by the cameras, the output image is populated using the computed mappings for each camera. The color is bi-linearly interpolated from the camera images.

\subsection{Cloud Coloring} \label{sec:cloud_coloring}

The Lidar cloud is colored online by assigning colors to each point as soon as a new scan cloud is received. For this, each point of the scan is projected into the latest camera images. This follows the same idea as the previously described projection method. The 3D projection surface is replaced by the 3D points retrieved from the Lidar scans.

Lidar scan points $\mathbf{x}_L$ are given relative to the Lidar sensor frame $L$. They are transformed to each camera frame using the extrinsic Lidar-camera calibration $\transform{T}{R}{L}$ and the extrinsic camera calibration $\transform{T}{C_i}{R}$ of each camera:

\begin{equation}
    \mathbf{x}_{c_i} = \transform{T}{C_i}{R} \,\transform{T}{R}{L} \mathbf{x}_L
\end{equation}

Using the intrinsic calibration, the points are projected into all camera images using (\ref{eq:camera_projection}). If a point is visible from multiple cameras, the observation closest to the respective image center is preferred. The color is interpolated bi-linearly. This process is repeated for each point of the scan and for each camera.

\section{RESULTS}

In this section, we highlight advantages of our proposed methods compared to traditional approaches in multiple use-cases. We use two different multi-camera systems for demonstration: the compact omnidirectional camera \textit{Insta360 Air} mounted on our tracked platform \textit{Asterix} and the quadrupedal robot \textit{Boston Dynamics Spot}.

\subsection{Insta360 Air}
The \textit{Insta360 Air} (Fig.~\ref{fig:insta360_air}) is a compact omnidirectional camera composed of two 210° fish-eye lenses on a spherical body with a diameter of 4 cm.
The camera is mounted on our highly mobile tracked robot \textit{Asterix}~\cite{schnaubelt_mechatronik_2021} (Fig.~\ref{fig:asterix}) and is positioned on top of the central frame for maximum visibility. By replacing our previous multi-camera system with an omnidirectional camera, we reduced system complexity and space requirements significantly while solving the same use-cases. Additionally, calibration effort is reduced to a single camera with two lenses.

\begin{figure}
    \centering
    \includegraphics[width=\linewidth]{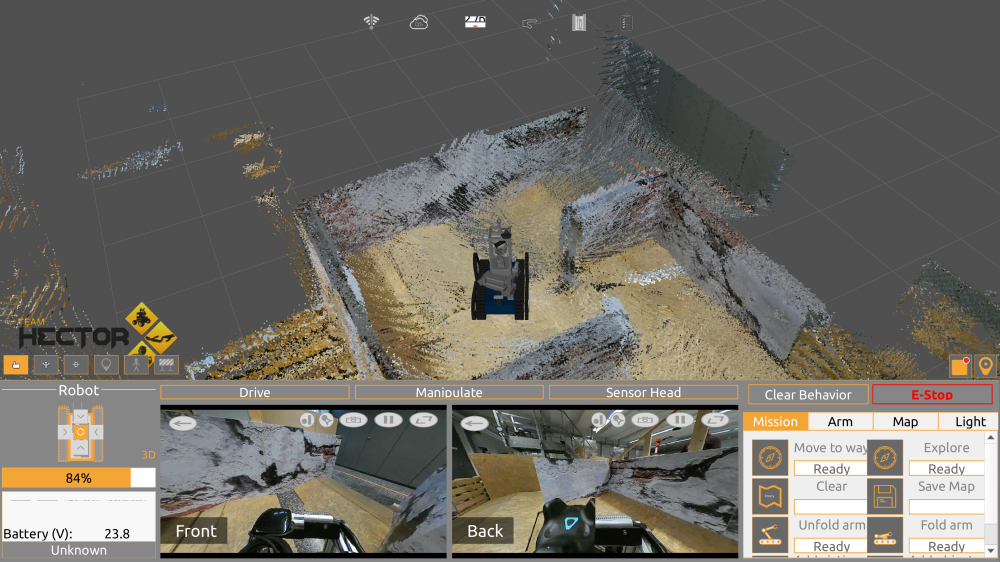}
    \caption{Operator User Interface. The main view visualizes the colorized point cloud in a virtual representation of the environment. Below the main view, projected forward and backward cameras are shown. Their viewing direction can be adjusted online with a joystick.}
    \label{fig:user_interface}
\end{figure}

In the following, we present the integration of our approach into a single comprehensive user interface (see Fig.~\ref{fig:user_interface}). It is composed of a virtual 3D scene in the main window and virtual views underneath.

\textit{Virtual views:} We use two pinhole projections for virtual forward and backward driving cameras, giving the operator the capability to quickly switch the driving direction while also providing additional situation awareness behind the robot as advised in~\cite{keyes2006camera}. The projections feature a high \ac{fov} of 130° for improved visibility to the sides. Using projection, we can achieve a high \ac{fov} without distortions as would be observable on a comparable fish-eye lens. To ensure accuracy of the projection and reduce stitching seams, the projection surface can be adapted to the standard operational viewing distance by the user.

Additionally, the operator can rotate and zoom the projections to inspect the robot's surroundings without restrictions. This functionality is realized by recomputing the projection based on joystick input. This effectively implements a virtual sensor head without the mechanical complexity and space requirements of an actuated pan-tilt unit. Additionally, we can track arbitrary frames on the robot or the environment, for instance a manipulator end-effector.

The projections are generated directly on the robot to reduce bandwidth requirements. Only images of regions of interest are transmitted instead of the high-resolution spherical source images.

Creating virtual projections allows us to use the camera for more than one purpose at a time. Simultaneous to its function as an operator view, we use the camera for 360$^\circ$ object detection. A panoramic view of the environment based on the Mercator projection (see Fig.~\ref{fig:image_projection_preview}) is generated which is used to detect hazmat signs in the environment.

\begin{figure}[tbp]

    \begin{subfigure}[t]{0.48\linewidth}
        \includegraphics[width=\linewidth]{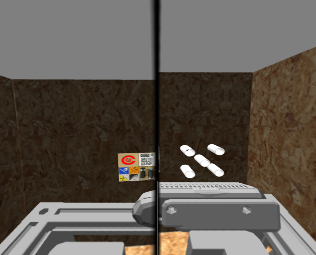}
        \caption{Vunder et al.}\label{fig:comparison_vunder_theirs}
    \end{subfigure}
    \begin{subfigure}[t]{0.48\linewidth}
        \includegraphics[width=\linewidth]{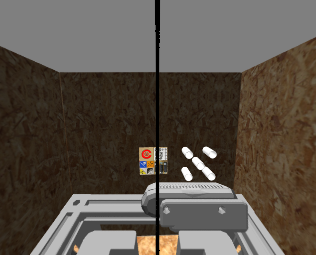}
        \caption{Ours}\label{fig:comparison_vunder_ours}
    \end{subfigure}
    \\
    \begin{subfigure}[t]{0.48\linewidth}
        \includegraphics[width=\linewidth]{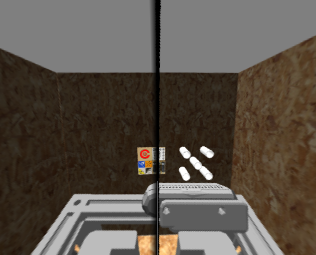}
        \caption{Vunder et al., corrected by our approach}\label{fig:comparison_vunder_theirs_corrected}
    \end{subfigure}%

    \caption{Comparison between Vunder et al.~\cite{vunder2018improved} and our approach on a simulated omnidirectional camera with two lenses showing an offset of $3^\circ$ around the optical axis. In (\subref{fig:comparison_vunder_theirs}), the wall shows a misalignment, which is not present in our approach (\subref{fig:comparison_vunder_ours}). By correcting the fish-eye images in a pre-processing step using our approach, the wall is aligned (\subref{fig:comparison_vunder_theirs_corrected}).}\label{fig:comparison_vunder}
\end{figure}

Next, we compare the pinhole projection in our approach to the one described by Vunder et al.~\cite{vunder2018improved} (Fig.~\ref{fig:comparison_vunder}). The approaches are fundamentally different as our approach creates pinhole projections directly as opposed to an intermediate mapping to a sphere. This intermediate step relies on the assumption that two ideal spherical images are used and the camera centers coincide. This assumption is rarely met on real cameras and leads to artifacts, as can be seen in the misalignment of the wall in Fig.~\ref{fig:comparison_vunder_theirs}. Our approach considers extrinsic and intrinsic calibration of each camera and, therefore, does not show these artifacts (Fig. \ref{fig:comparison_vunder_ours}). We can correct the projection using our approach by first computing an ideal spherical projection from the original fish-eye images (Fig.~\ref{fig:comparison_vunder_theirs_corrected}).

\begin{figure}[tbp]

    \begin{subfigure}[t]{0.48\linewidth}
        \includegraphics[width=\linewidth]{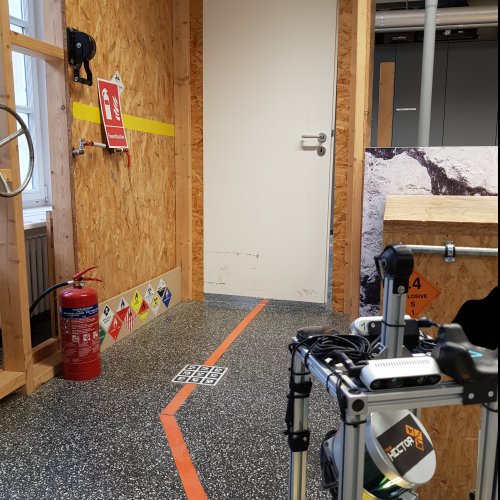}
        \caption{Reference photo}\label{fig:comparison_camera_view}
    \end{subfigure}%
    \begin{subfigure}[t]{0.48\linewidth}
        \includegraphics[width=\linewidth]{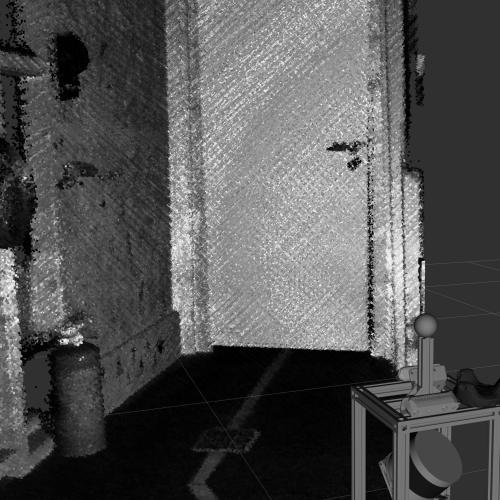}
        \caption{Colored by intensity}\label{fig:comparison_intensity}
    \end{subfigure}%
    \\
    \begin{subfigure}[t]{0.48\linewidth}
        \includegraphics[width=\linewidth]{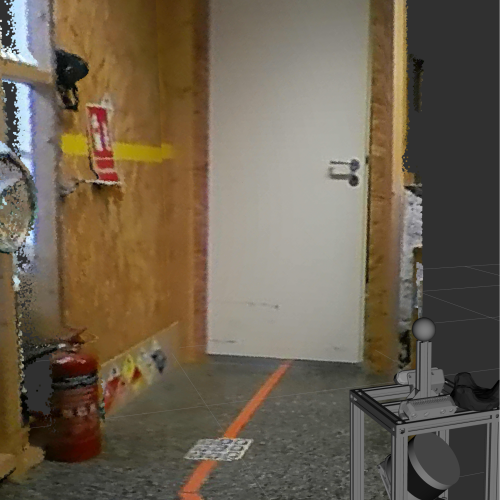}
        \caption{Proposed approach}\label{fig:comparison_colored}
    \end{subfigure}%
    \begin{subfigure}[t]{0.48\linewidth}
        \includegraphics[width=\linewidth]{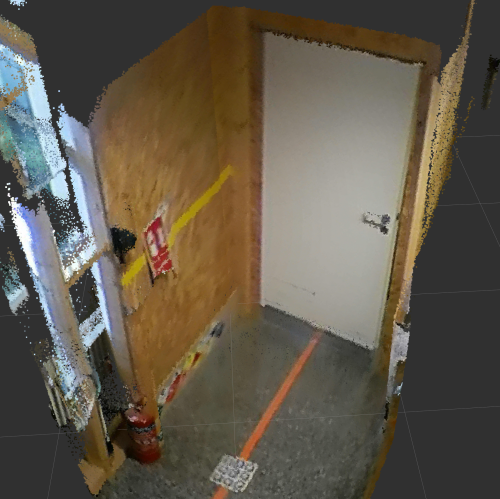}
        \caption{Different perspective}\label{fig:comparison_colored_second}
    \end{subfigure}%

    \caption{The colorized point cloud provides improved semantic understanding by making objects recognizable by color.}\label{fig:colored_point_cloud_comparison}
\end{figure}

\textit{3D view:} The robot is equipped with a rotating tilted \textit{Velodyne VLP-16} Lidar that generates a point cloud of the environment. The cloud is colorized online with the omnidirectional camera mounted above. Cloud points that are obstructed by robot parts are filtered to prevent parts of the robot itself being projected onto the environment.
A 3D scene is displayed to the operator above the virtual camera views (Fig.~\ref{fig:user_interface}). Following the conclusions in~\cite{norton2017analysis}, spatial awareness is increased by visualizing the colorized cloud in this virtual representation of the environment that allows the operator to freely change the viewpoint. Based on the geometric data of the Lidar alone, it is very challenging for a human to develop a semantic understanding of the robot's environment. Consider the comparison made in Fig.~\ref{fig:colored_point_cloud_comparison}. A reference photo of the scene is given in Fig.~\ref{fig:comparison_camera_view}. Fig.~\ref{fig:comparison_intensity} shows the Lidar data, viewed from the same angle and colored by return intensity. While this view already allows us to identify obstacles and some objects, the white door, fire extinguisher and signs on the left are easier to identify in the colored cloud (Fig.~\ref{fig:comparison_colored}). Therefore, this approach combines spatial awareness through precise Lidar clouds with semantic information provided by colored camera data. Additionally to the benefits for teleoperation, the color information can also be leveraged for improved automatic 3D object detection.

\subsection{Boston Dynamics Spot}

\begin{figure}[tbp]
    \begin{subfigure}[t]{\linewidth}
        \begin{minipage}[b]{0.53\linewidth}
            \includegraphics[width=0.48\linewidth]{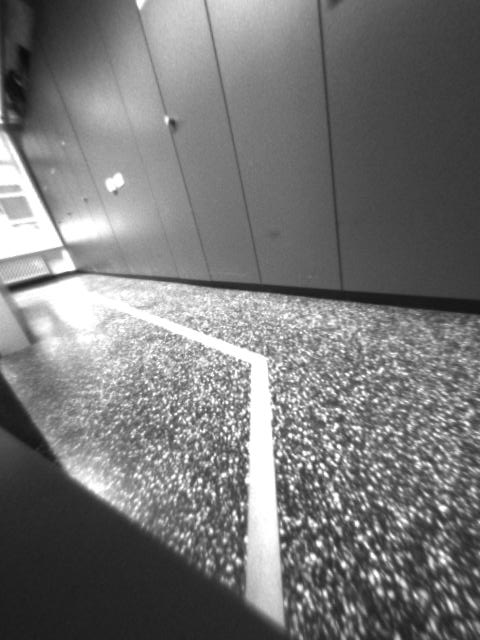}
            \includegraphics[width=0.48\linewidth]{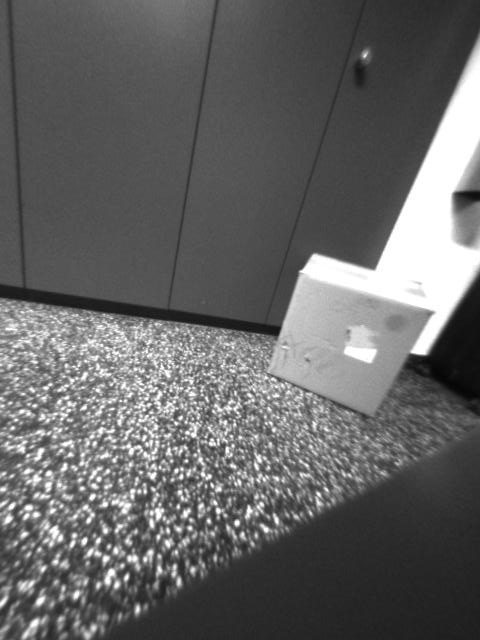}
        \end{minipage}
        \begin{minipage}[b]{0.456\linewidth}
            \includegraphics[width=0.48\linewidth]{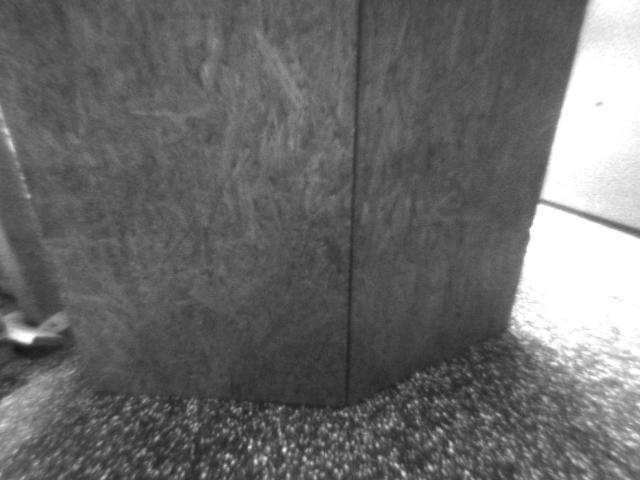}
            \includegraphics[width=0.48\linewidth]{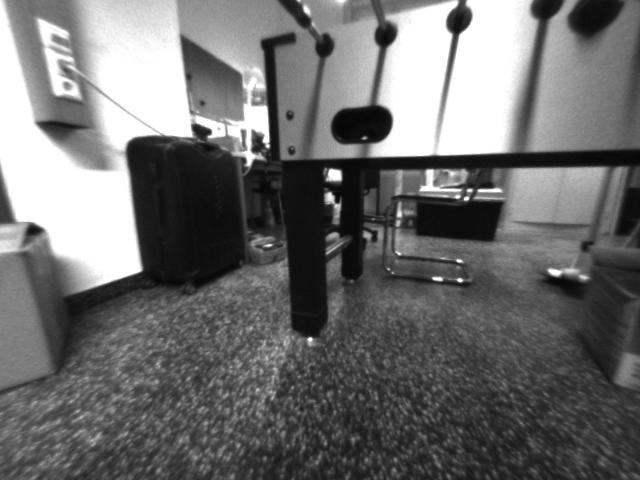}

            \vspace{3px}

            \includegraphics[width=0.48\linewidth]{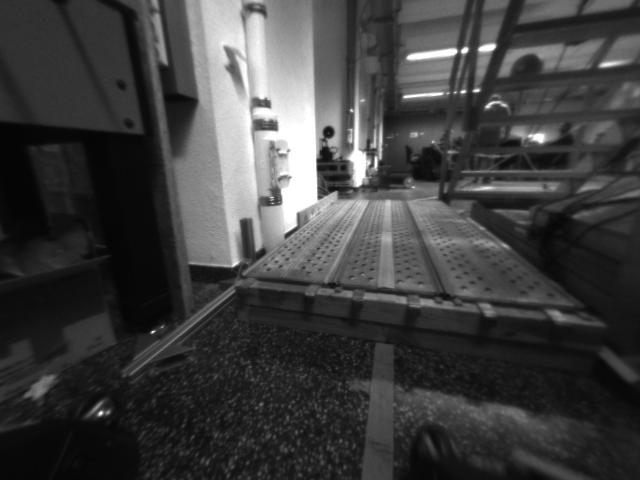}
        \end{minipage}
        \caption{Spot camera images} \label{fig:spot_cams}
    \end{subfigure}
    \vspace{5px}

    \begin{subfigure}[b]{0.58\linewidth}
        \includegraphics[width=\linewidth]{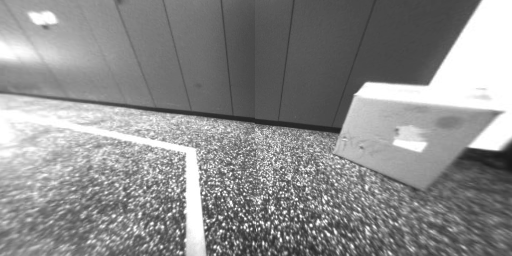}
        \caption{Virtual front view} \label{fig:spot_frontfused}
    \end{subfigure}
    \begin{subfigure}[b]{0.4\linewidth}
        \includegraphics[width=\linewidth]{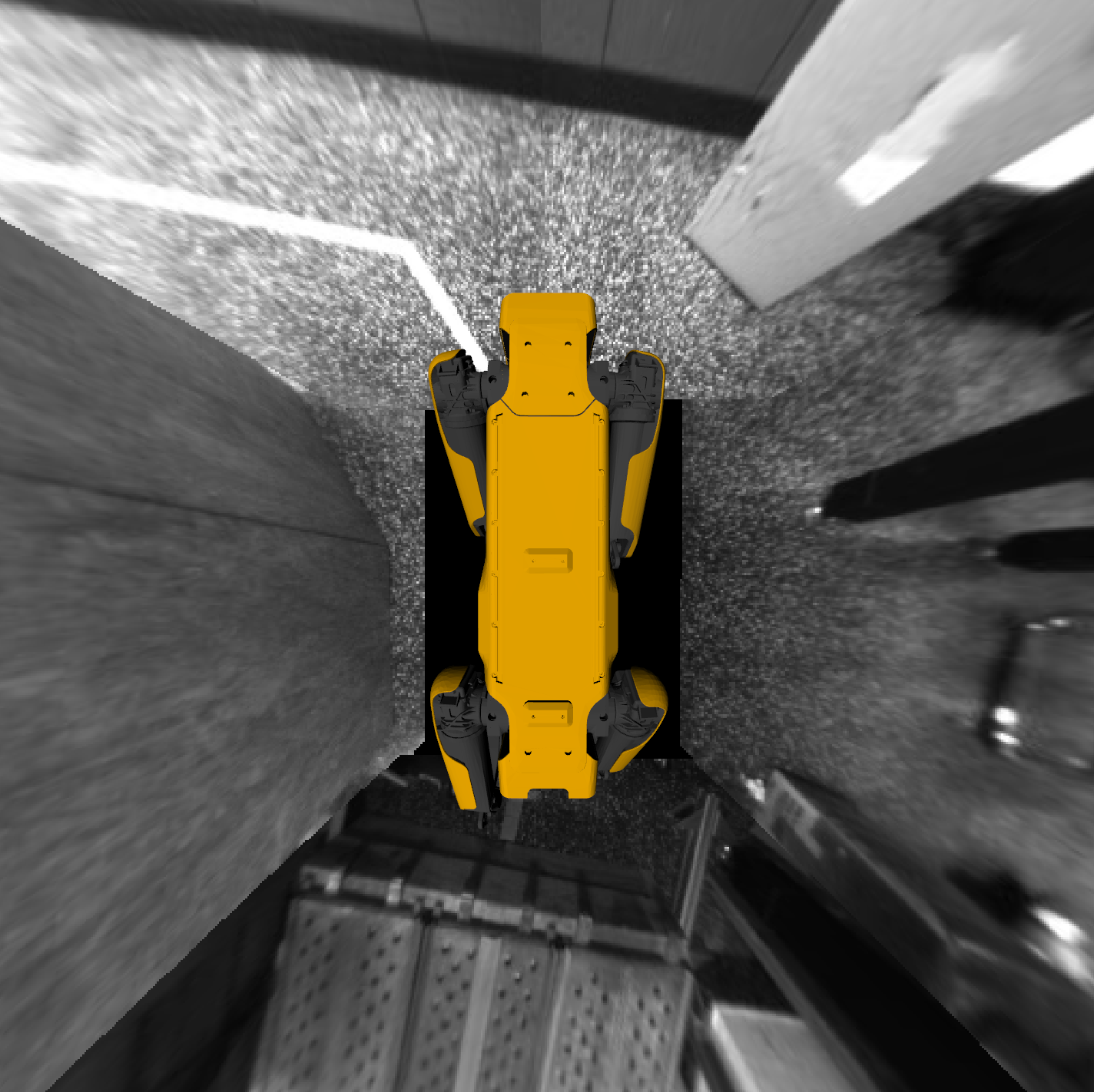}
        \caption{Virtual bird's-eye} \label{fig:spot_birdseye}
    \end{subfigure}%

    \caption{The five source camera images of Spot can be seen in (\subref{fig:spot_cams}). They are used to compute a forward view (\subref{fig:spot_frontfused}) and a bird's eye view with overlayed robot model (\subref{fig:spot_birdseye}) for additional situation awareness.}\label{fig:spot_projections}
\end{figure}

The \textit{Boston Dynamics Spot} is a quadrupedal robot equipped with a camera system composed of five grayscale pinhole cameras for omnidirectional vision.
Compared to the previous omnidirectional camera application example, this camera configuration is significantly different and highlights the generality of the proposed approach.
The five source camera images can be seen in Fig.~\ref{fig:spot_cams}. The robot does not possess a forward-facing camera. Therefore, we use a perspective projection to fuse the two angled front cameras to  generate a view suitable for teleoperation (Fig.~\ref{fig:spot_frontfused}). With a horizontal \ac{fov} of 130°, the resulting projection provides better peripheral vision than a single camera.
As a legged platform, \textit{Spot} can move in any direction. Switching between cameras to be aware of all movement directions simultaneously is demanding and error-prone~\cite{chen2007human}.
We address this issue by generating a virtual bird's-eye view (Fig.~\ref{fig:spot_birdseye}) overlayed with the rendered robot model, showing the robot and its environment in a single image as recommended by~\cite{keyes2006camera}. This view is achieved with a perspective projection by positioning the virtual camera above the robot and the projection surface on ground level. Compared to \cite{sato2013spatio}, the projection is generated directly without any intermediate steps. While this visualization increases situation awareness in close proximity to the robot, the operator has to be aware of the fact that objects are distorted, because they are higher than the ground.

\subsection{Performance Evaluation}
The performance has been evaluated on an \textit{Intel i7-4710HQ@2.50GHz} mobile CPU and the \textit{Insta360 Air} camera. We tested different projections and target resolutions. The required computation time for the two processing steps is shown in Table~\ref{tab:performance_evaluation}. The projection mapping has to be generated only when the projection changes, e.g. on startup or when the view is rotated. The computation time scales linearly with the target resolution. This step is rather costly compared to the map operation which applies the mapping to the camera images to generate a new projection frame. Even with high target resolutions, a real-time application is possible. The cloud colorizing performance has been tested with the \textit{VLP-16} Lidar. In our setup it produces about 18000 points per scan with a rate of 10 Hz. Processing of a single scan took 57.72 ms on average, allowing for real-time usage.

\begin{table}[tbp]
    \centering
    \caption{Image Projection Performance}
    \label{tab:performance_evaluation}
    \begin{tabular}{c|c|c|c}
               &            & Projection & Map   \\
    Projection & Resolution & mapping (ms) & operation (ms) \\
    \hline
    Perspective &  512x256 & 23.78 & 12.74 \\
    Perspective &  1024x512 & 86.27 & 12.5 \\
    Perspective &  2048x1024 & 334.62 & 23.36 \\
    Mercator & 1024x512 & 141.67 & 12.33 \\
    Spherical &  512x512 & 92.26 & 10.9 \\
    \end{tabular}
\end{table}

\section{CONCLUSION}

We presented a novel flexible framework for providing virtual projections to remote robot operators based on multiple cameras anywhere on the robot as well as a complementary approach to colorize Lidar scans from image data. The software implementation of both approaches is available as open-source ROS packages. The combination of both approaches in a comprehensive user interface allows for significantly improved situation awareness during teleoperation.

As future work, we are planning to evaluate our approach on omnidirectional cameras with increased resolution.
Furthermore, we expect GPU acceleration to significantly speed up computation times as the structure of data processing is well suited for parallelization.

\bibliographystyle{IEEEtran}
\bibliography{IEEEabrv, literature}

\end{document}